\documentclass[letterpaper, 10 pt, conference, top=30pt]{ieeeconf}  

\IEEEoverridecommandlockouts                              

\usepackage[utf8]{inputenc}
\usepackage{graphicx}
\usepackage{subcaption}
\usepackage{booktabs}
\usepackage{amsmath} 
\usepackage{amssymb}  
\usepackage{float}

\title{\LARGE \bf
Temporal Feature Networks for CNN based Object Detection
}

\author{Michael Weber$^{1}$, Tassilo Wald$^{1}$ and J. Marius Zöllner$^{1}$
\thanks{$^{1}$ When this work was done, all authors  were  with FZI Research  Center  for  Information  Technology, 76131 Karlsruhe, Germany
        {\tt\small \{michael.weber, wald, zoellner\}@fzi.de}}%
}

\begin{document}

\maketitle
\thispagestyle{empty}
\pagestyle{empty}

\begin{abstract}
For reliable environment perception, the use of temporal information is essential in some situations.
Especially for object detection, sometimes a situation can only be understood in the right perspective through temporal information.
Since image-based object detectors are currently based almost exclusively on CNN architectures, an extension of their feature extraction with temporal features seems promising.

Within this work we investigate different architectural components for a CNN-based temporal information extraction. We present a Temporal Feature Network which is based on the insights gained from our architectural investigations. This network is trained from scratch without any ImageNet information based pre-training as these images are not available with temporal information. The object detector based on this network is evaluated against the non-temporal counterpart as baseline and achieves competitive results in an evaluation on the KITTI object detection dataset.

\end{abstract}

\section{INTRODUCTION}
In autonomous driving, the environment perception still heavily relies on image based perception methods as there is a lot of relevant information in the environment that cannot be captured by other sensors e.g. the state of a traffic light~\cite{weber18} or brake lights.
Also the use of temporal information might be essential for a correct understanding of the actual surrounding scene. Objects might be partly occluded in single frames and in general information about the movement and  velocity of an object has primarily to be obtained from temporal information. 
In a few areas of perceptions for autonomous driving like ego-motion estimation~\cite{weber17}, there have already been used CNN-based temporal information extractors. 
Also in the area of directly steering a vehicle based on a camera input -- the so called End-to-End driving~\cite{hubschneider17} -- CNN-based temporal feature extraction approaches have been successfully applied.

Current object detectors are usually based on well-known CNN backbones, so called feature extractors. RetinaNet~\cite{lin17a} e.g. is based on ResNet~\cite{he16}, the older SSD~\cite{liu16} is based on a VGG16~\cite{simonyan14} architecture and YOLOv2~\cite{redmon17} implements an architecture based on GoogLeNet~\cite{szegedy2016rethinking} because it is faster than VGG16 and requires less memory. It is obvious to adapt and train an architecture based on the known backbones for image sequences.
These common feature extraction networks are usually trained initially on the ImageNet classification dataset~\cite{deng09} for a long time on large server clusters in order to obtain optimal weights for the convolution filters.

\begin{figure}[ht]
\centering
	\includegraphics[width=.45\textwidth]{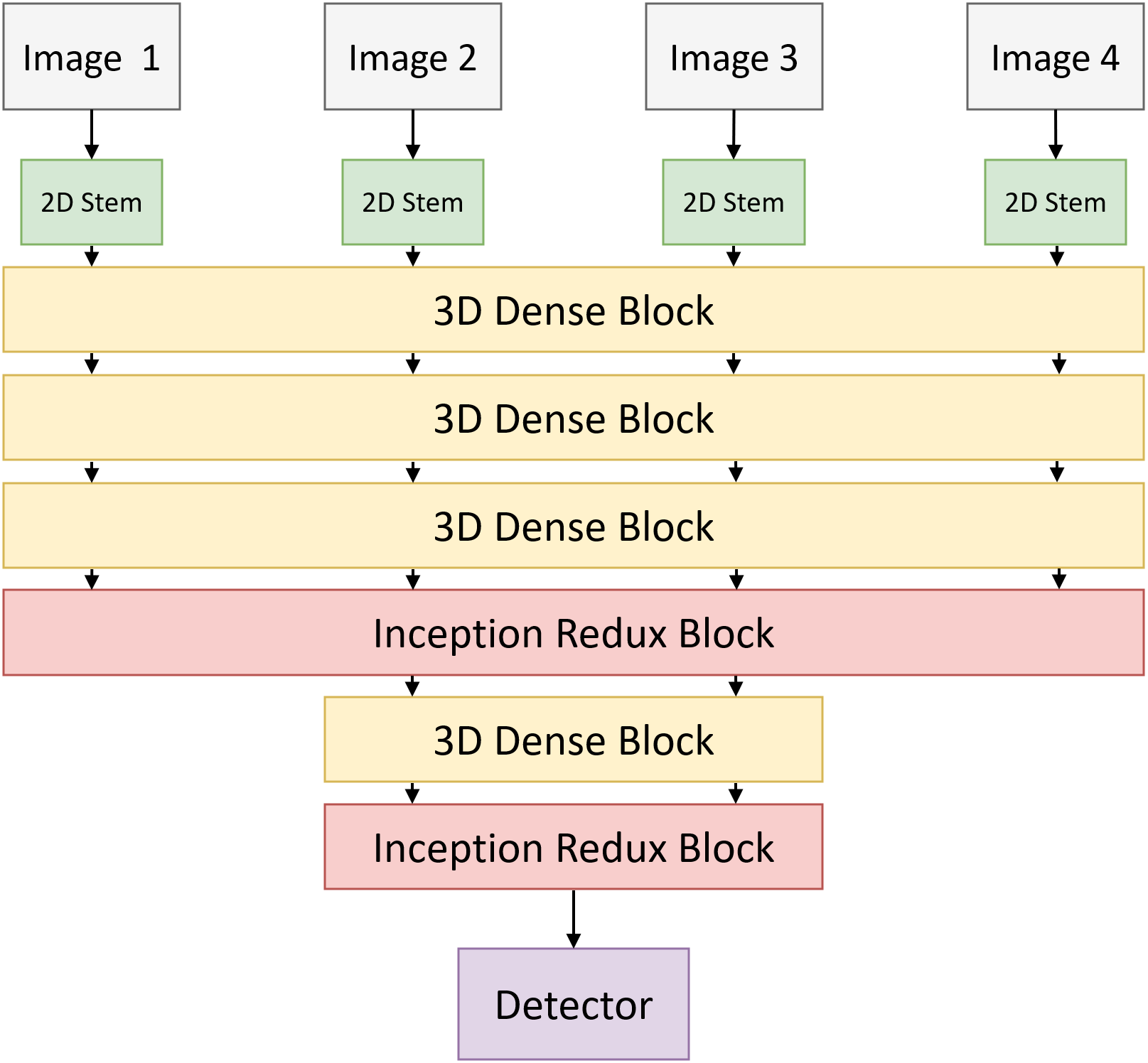}
	\caption{The architecture of the Temporal Feature Network based object detector. The width of the blocks corresponds to their temporal depth. Each dense block
	is succeeded by a transition block. The red blocks are Temporal Fusion Blocks.}
	\label{temp_fus_arch_img}
\end{figure}

However, for temporal information extraction this procedure has a considerable disadvantage: there exists no direct procedure to obtain pre-trained weights for temporal convolution  kernels.
For CNN-based temporal information extraction usually an architecture based on the slow fusion principle~\cite{karpathy2014large} is applied.
The convolutions within this architecture are so-called 3D convolutions which are extended by one dimension.
This prevents the straightforward use of pre-trained weights from classical and well-known feature extraction networks.

With the Deeply Supervised Object Detector (DSOD)~\cite{Shen_2017_ICCV} a object detector emerged that promises a training from scratch without using ImageNet and with relatively moderate training efforts.
Since this seems to be a promising step towards training temporal object detectors from scratch, we are conducting intensive architecture studies based on this detector.
As a result, we introduce the Temporal Feature Network as backbone for an image-based object detector capable of gaining advantage from temporal information present in image sequences. 

\section{RELATED WORK}

\subsection{Object Detection}
Image-based CNN object detectors were one of the first use-cases in environment perception for autonomous driving. While the so-called two-stage detectors like Region-CNN~\cite{girshick14} and its descendants only played a minor role in autonomous driving due to their long and unpredictable runtime, one-stage detectors like the YOLO family (\cite{redmon16}, \cite{redmon17}, \cite{bochkovskiy20}, \cite{redmon18b}), SSD~\cite{liu16}, RetinaNet~\cite{lin17a} and already OverFeat~\cite{sermanet13} made their way into autonomous driving functions like~\cite{huval15}.
Recently also approaches for 3D object detection in the autonomous driving domain emerged \cite{weber19}, \cite{gahlert19}.

All of these approaches have in common that their feature extraction heavily relies on fine-tuning a feature extraction backbone like ResNet~\cite{he16}, VGG~\cite{simonyan14} or GoogLeNet~\cite{szegedy2016rethinking} that is pre-trained on the ImageNet dataset~\cite{deng09}.
With the Deeply Supervised Object Detector (DSOD) \cite{Shen_2017_ICCV}, there exists only one common object detector which can be trained without fine-tuning these pre-trained weights.
The authors claim that deep supervision is essential for training from scratch.
A similar claim, for the necessity of deep supervision, is done in \cite{jegou2017one}.

In difference to the other mentioned approaches, Densely Connected Convolutional Networks (DenseNets) \cite{huang2017densely} use dense blocks in which the output of each convolutional layer is the input of each following convolutional layer.

\subsection{Temporal Information Fusion}
In \cite{karpathy2014large} different principles to fuse information over temporal dimension in a CNN are presented and used for video classification.
In~\cite{weber17}, visual ego-motion estimation is performed on the basis of similar network architectures.

For the object detection task, in the last few years LiDAR data has been used in conjunction with camera images. However, there has been little research into using temporal information from multiple images for object detection. \cite{zhu2017flow} presents an approach called Dense Feature Aggregation which aggregates feature maps and \cite{simonyan14b} introduced a two-stream network for classification where the visual information and the temporal information are processed separately in parallel streams.

\section{TEMPORAL FEATURE NETWORK}

We propose a neural network architecture (see Fig.~\ref{temp_fus_arch_img}) which is inspired by DSOD~\cite{Shen_2017_ICCV} and enhanced with temporal information fusion.
It takes a series of subsequent images, called a sequence, as input and predicts objects on the last image.
The temporal fusion architecture follows the slow fusion design paradigm which has proven to work well for video classification~\cite{karpathy2014large} and visual ego-motion estimation~\cite{weber17}.

The input sequence is forwarded through the stem of the model, which is identical to the 2D DSOD object detector but reproduced to process each input image.
The stems of all images have shared weights to create a unified initial feature extractor.
After the stems, the input is passed to a 3D dense block. The architecture of dense blocks is identical to that of DenseNet~\cite{huang2017densely}, but all convolutional layers use 3D filters. Each k$\times$k convolutional filter is inflated to k$\times$k$\times$k while also inflating the padding to keep the size of dimensions constant. The growth factor of all dense blocks is 48.

After each dense block, a transition block, similar to DSOD transition block, is added. It first applies a 1$\times$1$\times$1 convolution for bottleneck purposes. Then a 2D max-pool is used to halve the height and width. The depth is not reduced, so its dimension remains unchanged. Similar to DSOD, the last transition block in the architecture does not reduce the resolution but only uses its bottleneck layer.

To reduce the temporal depth, different temporal fusion blocks are introduced and evaluated. With Late Slow Fusion and Early Slow Fusion, two different ways of integrating these temporal fusion blocks into the DSOD architecture are presented.

\begin{figure}[h]
	\includegraphics[width=.485\textwidth]{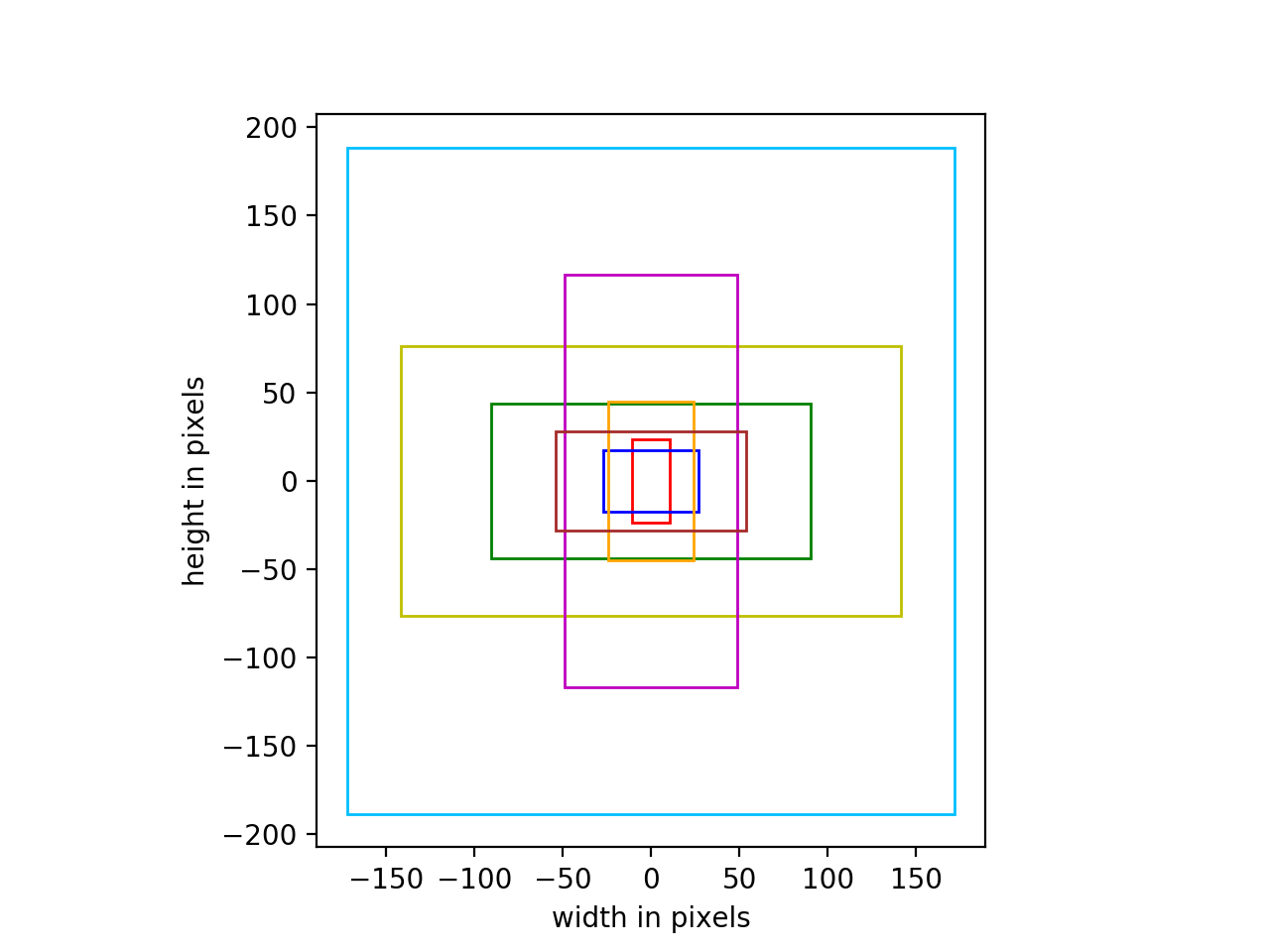}
	\caption{Shapes of the eight created anchor boxes.}
	\label{anchor_boxes}
\end{figure}


We have deliberately defined anchor boxes to have high IoU scores with the bounding boxes in the training set. For this, we have analyzed all objects in the our training dataset and clustered them using k-means with IoU as a distance metric. As a result we have obtained eight anchor boxes (see Fig.~\ref{anchor_boxes}), which comprise various box forms to detect both pedestrians (upright vertical rectangles) and cars (horizontal rectangles).

After the final temporal fusion block, the temporal dimension of 1 is collapsed. Then a final 2D convolutional layer is added, which is used as the detector. It uses 72 1$\times$1$\times$1 filters without padding. Each pixel in this feature map predicts a bounding box for each anchor. Since 8 anchor boxes are used, this results in $32 \cdot 18 \cdot 8 = 4608$ predicted bounding boxes.

Similar to \cite{redmon17}, the offsets to the anchor box $t_x,t_y,t_w, t_h$ as well as a confidence score $t_{conf}$ are predicted for each bounding box. Additionally, two values for class certainty ($t_{car}, t_{ped}$) are also predicted for each bounding box. 


\begin{figure}[H]
\centering
	\includegraphics[width=.43\textwidth]{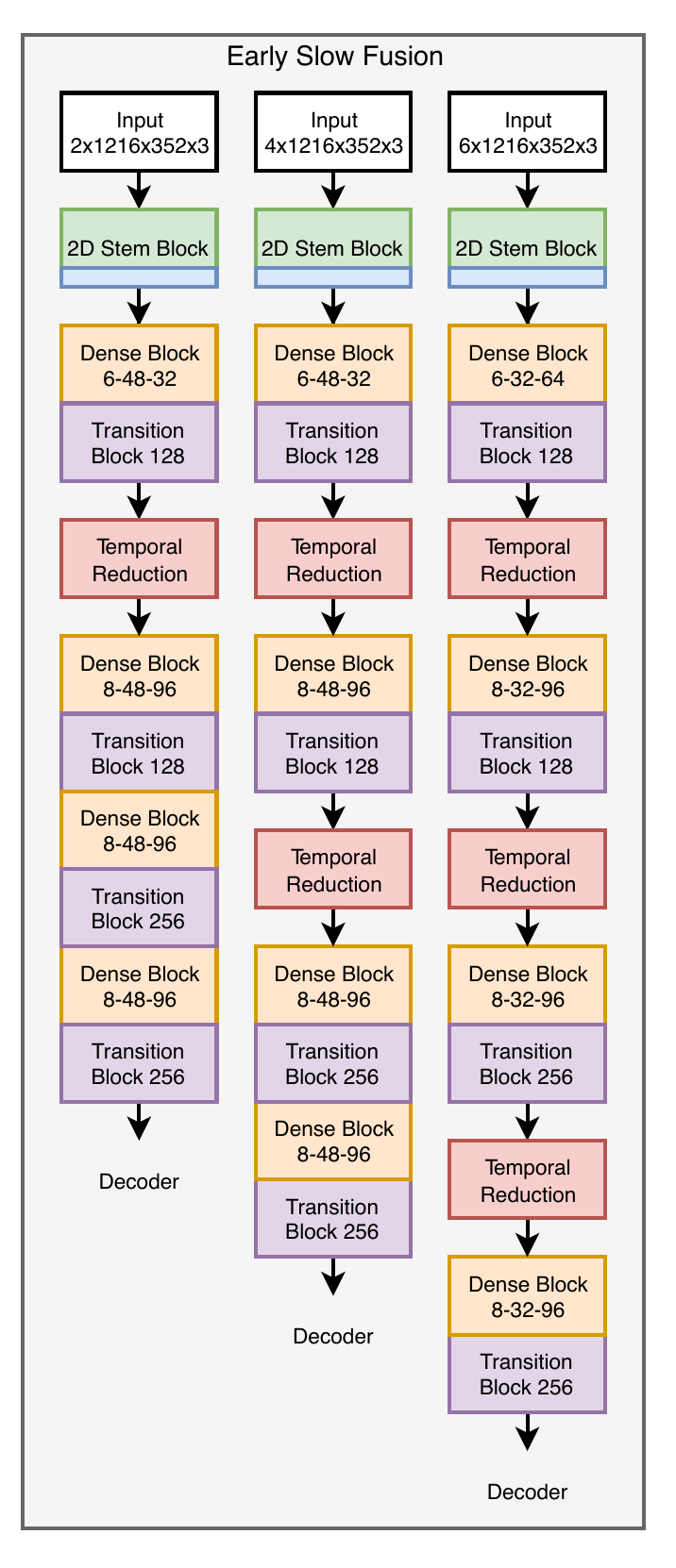}
	\caption{Early Slow Fusion architecture for different input sequence lengths. Depicted is the position of the temporal reduction blocks (red) within the encoder network for an input sequence length of two (left), four (middle) and six (right).}
	\label{fig:arch_early}
\end{figure}

Because of high memory constraints of the feature maps, the proposed network architecture contains only 5 million parameters, which is less than comparable networks (Resnet50 with 23 million and original DSOD with 14 million parameters).
In the following sections, some aspects are examined further in greater detail.

\subsection{Temporal Fusion Blocks}
The reduction of temporal depth across the entire feature extraction layers is the central characteristic of slow fusion networks~\cite{karpathy2014large}. 
Unlike the early fusion and late fusion design principles, however, the greatest creative leeway remains available for this principle.
Inspired by the other design principle, we developed two different architectures for our Temporal Feature Network.

\textbf{Early Slow Fusion Network:} The Early Slow Fusion (ESF) is inspired by the early fusion principle~\cite{karpathy2014large} and shifts temporal depth reduction towards the earlier layers of the feature extraction or encoder network. The architecture for different input stream sizes is shown in Fig.~\ref{fig:arch_early}.

\begin{figure}[H]
\centering
	\includegraphics[width=.45\textwidth]{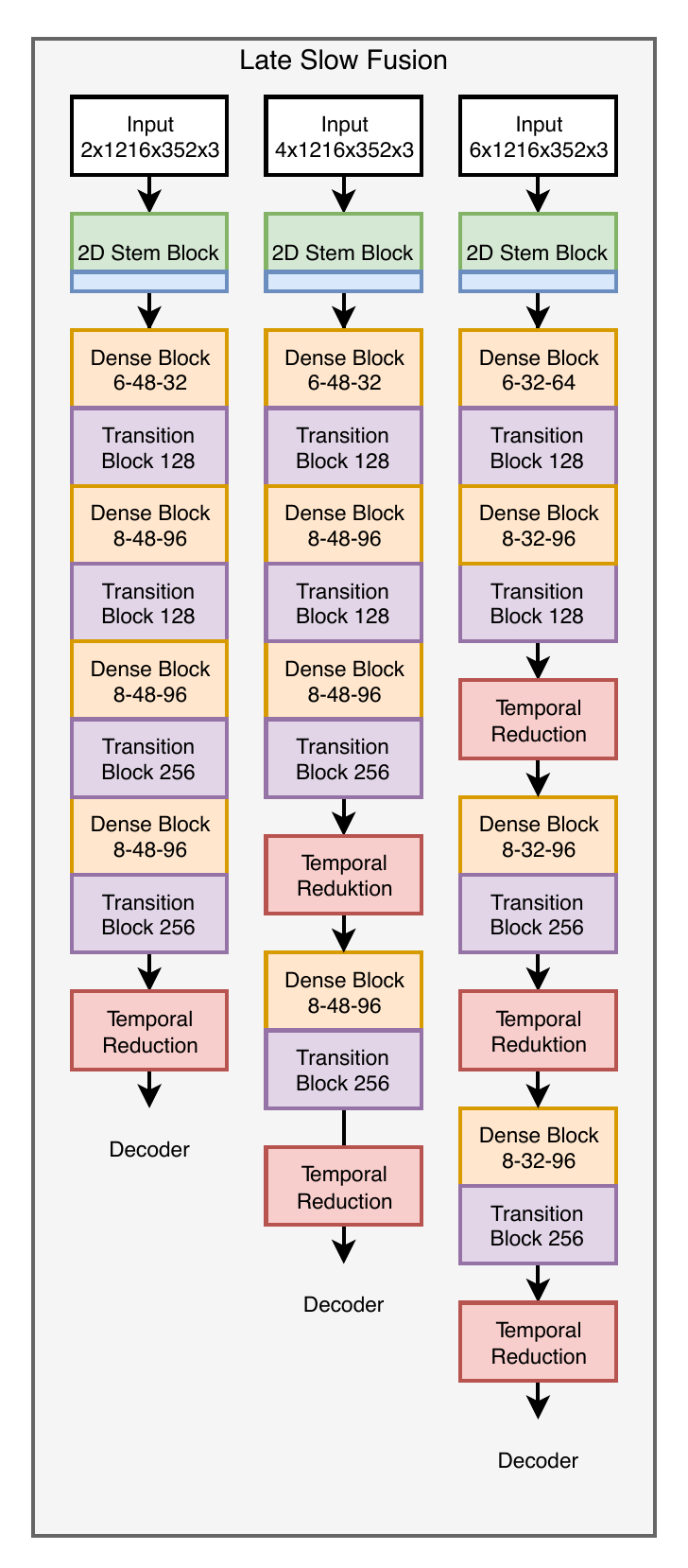}
	\caption{Late Slow Fusion architecture for different input sequence lengths. Depicted is the position of the temporal reduction blocks (red) within the encoder network for an input sequence length of two (left), four (middle) and six (right).}
	\label{fig:arch_late}
\end{figure}

\textbf{Late Slow Fusion Network:} The Late Slow Fusion (LSF) is inspired by the late fusion principle~\cite{karpathy2014large} and therefore was designed to shift the temporal depth reduction towards the later layers of the encoder network. The architecture for different input stream sizes is shown in Fig.~\ref{fig:arch_late}.

After introducing the high level architectures, now the layers in which the reduction of the temporal depth finally takes place shall be looked at in more detail.
In CNN architectures, the reduction of the size of the feature maps usually takes place in pooling layers with step size two, which leads to a halving of the size. In convolutional layers, the size of the feature maps is kept constant by padding. Due to the small temporal depth of two to six images, halving the temporal depth is not mandatory.
Therefore, besides different pooling layers it is also investigated which filter kernels are optimal to reduce the temporal information by convolutions. For this purpose, a number of different temporal reduction layers are evaluated:

$\mathbf{T \times 1 \times 1 }$\textbf{ Conv:} A layer of multiple $T \times 1 \times 1 $ filter kernels is the simplest way to fuse temporal information. However, this filter kernel cannot represent a spatial motion component - only a temporal progression. The need for parameters is therefore very low.

$\mathbf{T \times 3 \times 3 }$\textbf{ Conv:} Offers the possibility to represent spatial motion information. Compared to the $T \times 1 \times 1 $ filter kernel, the requirement of parameters is 9 times larger.

$\mathbf{T \times 5 \times 5 }$\textbf{ Conv:} Provides a larger receptive field than the $T \times 3 \times 3 $ kernel and can therefore resolve spatial motion more fine-grained. However, the parameter number increases quadratically and is 25 times larger than for the $T \times 1 \times 1 $ filter.

\textbf{3D Inception v1:} Is based on parallelizing a $1 \times 1$, $3 \times 3$, and $5 \times 5$ convolution as well as a $3 \times 3$ max pooling operation. All filter kernels of these layers are extended by T in the temporal dimension so that they are $T \times k \times k$ in size. The number of parameters increases significantly as a result.

\textbf{3D Inception v2:} Consists, like 3D Inception v1, of several parallel convolution layers and a parallel pooling layer. The difference in this case, however, is that the filter kernels are split into the individual dimensions. Thus, a $3 \times 3$ convolution becomes a $1 \times 3$ convolution followed by a $3 \times 1$ convolution. To simulate a $5 \times 5$ convolution, two $3 \times 3$ convolutions are used to obtain the identical receptive field of the $5 \times 5$ convolution. This leads to a strong reduction of trainable parameters. 

$\mathbf{T \times 5 \times 5 }$\textbf{ Max Pool:} Extracts the maximum values of the feature map over time in the corresponding spatial frame. A mapping of motion information is not possible with this block. There are also no trainable parameters necessary as it is a pooling operation.

$\mathbf{T \times 5 \times 5 }$\textbf{ Mean Pool:} Determines the mean value of the feature map over the temporal course in the associated spatial frame. Analogous to the max pooling block, mapping of motion information is not possible and no trainable parameters are needed.

\subsection{Temporal Padding}
Since the reduction of temporal depth is already entirely performed within the temporal fusion blocks, it must remain constant within the other blocks of the network. 
This can be achieved with two different architectural choices:

\textbf{Non-temporal kernels:} In this case the convolution kernels outside the temporal fusion blocks have a temporal depth of 1 ($1 \times k \times k$), therefore the temporal depth of the feature maps remains unchanged. 

\textbf{Temporal kernels + padding:} For this choice, the temporal depth for the convolution kernels is greater than 1 which would result in a overall reduction of temporal depth.
To prevent this, temporal padding is added. As padding might add false information to the network information, a possible negative influence on network performance must be carefully monitored.
Since the majority of the convolutions are located within the Dense Blocks, essentially the Dense Blocks are extended from 2D to 3D with temporal padding.

\subsection{Temporal Horizon}
One of the key design parameters for our Temporal Feature Network is the size of the temporal horizon, the network is allowed to look back.
This temporal horizon is composed by the length of the input sequence and the distance of the single frames.

For obvious reasons, the minimum length of the input sequence is larger than one. But also the maximum length of the image sequence that can be fed into the network is limited. An increasing number of images will provide the network with more information, but the size of the feature maps also increases. Feature maps and parameters share the memory of the GPU. If the memory capacity is exceeded, training is impossible. Additionally, as the feature map size increases, the number of floating-point operations (FLOPS) in the network increases, greatly increasing the inference and training time. Therefore, it is imperative to find the minimum number of images that achieve maximum performance improvement. Through empirical testing, six images have emerged as the upper bound for architectures with sufficient layer depth and parameter number. Therefore, architectures that have between 2 and 6 images as temporal depth are investigated to find the most beneficial ratio of performance to inference time.

\newpage

\begin{figure}[ht]
\centering
	\includegraphics[width=.485\textwidth]{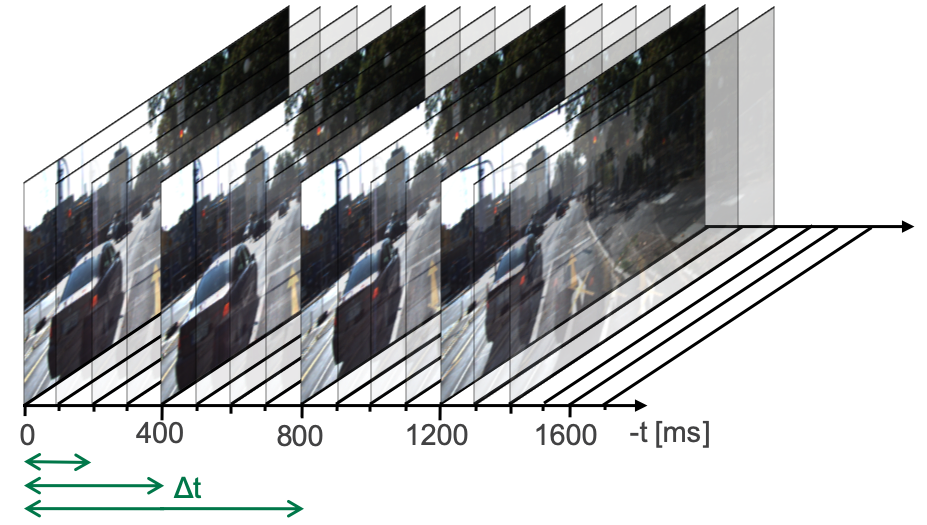}
	\caption{Selection of single images for the input sequence based on their temporal distance. The images here are selected with a temporal distance $\Delta t$ of 400~ms. All optional $\Delta t$ values that are investigated within this work are depicted as arrows (200~ms, 400~ms, 800~ms).}
	\label{fig:temp_dist_image}
\end{figure}

The images the input sequence is composed of can be selected according to the two different distance measures temporal distance and spatial distance.
The selection based on the spatial distance means that given the vehicle speed, the distance traveled is calculated, which selects the frames to be used.
This method can lead to problems if the sampling rate of the camera becomes too low for the driven speed.
In particular, however, the major problems arise when the vehicle speed is too low, since the time intervals between the individual images can become very large and therefore the temporal context might completely vanish.
Especially because of this a selection of images based on temporal distance $\Delta t$ is proposed. As shown in Fig.~\ref{fig:temp_dist_image} the input sequence images are selected based on the time elapsed between the acquisition of the individual images.
Experiments for finding an optimal $\Delta t$ are shown in the Evaluation section.

\section{EVALUATION}

\subsection{Dataset}
All experiments for this work have been conducted on the well known KITTI Object Detection dataset~\cite{geiger12}. The dataset contains 7481 images for training purposes at a resolution of $1242 \times 352 \times 3$ pixels. Since no validation dataset is available and multiple evaluation runs on the official test set are not permitted, the images in the training dataset are split for our experiments. 6000 images are used for training and 1481 for validation purposes, this corresponds to a split of 80:20.
The remaining training set still contains 40,570 objects.
For augmentation purposes, geometric augmentations horizontal flip and translation are applied. Also the texture augmentation techniques brightness, contrast, saturation and color jitter are applied. All techniques have been applied according to~\cite{romera18}.

\subsection{Training Details}
All experiments presented within this section have been performed on a single NVIDIA GEFORCE GTX 1080 Ti GPU with 11 GB RAM.
The RAM limitation results in a maximum batch size of 8 for our experiments. 
Each training has been run for 150 epochs -- given the batch size of 8 and the training set size of 6,000 this is equivalent to 129,000 iterations.
As optimizer, Adam~\cite{kingma2014adam} has been used with an initial learning rate of 0.001 and a continuous learning rate decay. 
The continuous decay reduces the learning rate by 75~\% every 50 epochs.

To avoid the generation of identical batches over several epochs, a shuffle buffer is used which contains 50 samples of the training set. Individual samples are drawn from this buffer in an equally distributed manner. The small size of the buffer is not an issue at this point since the images are already in a random order.

For evaluation purposes, the Average Precision measure from PASCAL VOC~\cite{everingham10} as also used in the KITTI object detection challenge~\cite{geiger12} has been used with one small adaption: As the results for some modifications might differ only marginally, the original 11-point approximation has been extended to a 101-point approximation. 

\subsection{Temporal Horizon}
In order to determine the best temporal horizon for the network input sequence, different experiments have been conducted to find the best number of images and their optimal temporal distance.
To find a lower bound on the input sequence length, the absolute performance of architectures with different sequence lengths is compared. The results are shown in Table~\ref{tab:temp_number_frames}. For the ESF architectures, the performance is very similar for two and four input images, the performance for simple and medium boxes increases slightly, while heavy boxes remain almost constant. The performance of the LSF architecture, on the other hand, increases significantly with a larger number of images for all object difficulties. Therefore, since both LSF and ESF benefit by increasing the number of input images, the number of input images is increased to four.
The results of the architectures with six images have a lower performance, since here the number of parameters was reduced and earlier a convolution with stride two had to be performed to reduce the memory consumption of the model. 

\begin{table}
    \centering
    \begin{tabular}{l@{ }cccc}
        \toprule
		Architecture & \#Images & AP Easy & AP Medium & AP Hard\\
		\midrule
		Late Slow Fusion & 2 & 0.841 & 0.827 & 0.742 \\
		Late Slow Fusion & \textbf{4} & \textbf{0.932} & \textbf{0.887} & \textbf{0.798} \\
		Late Slow Fusion & 6 & 0.835 & 0.816 & 0.731 \\
		\midrule
		Early Slow Fusion & 2 & 0.865 & 0.852 & 0.775  \\
		Early Slow Fusion & \textbf{4} & \textbf{0.870} & \textbf{0.856} & \textbf{0.776} \\
		Early Slow Fusion & 6 & 0.861 & 0.826 & 0.750 \\
		\bottomrule 
	\end{tabular}
\caption{Influence number of frames used for temporal fusion.}
 \label{tab:temp_number_frames}
\end{table}

Based on those results for the optimal input sequence length and the fact that the Late Slow Fusion clearly outperformed the Early Slow Fusion, the LSF and an input sequence length of 4 is used for the following experiment for finding the optimal temporal distance.
The results for this experiment are shown in Table~\ref{tab:temp_dist_table}.

\begin{table}
    \centering
    \begin{tabular}{lccc}
        \toprule
		Architecture & Temp. Distance & \#Images &  mAP\\
		\midrule
		Late Slow Fusion & 200 & 4 & \textbf{0.892}  \\
		Late Slow Fusion & 400 & 4 & \textbf{0.893} \\
		Late Slow Fusion & 800 & 4 & 0.863 \\
		\bottomrule 
	\end{tabular}
\caption{Influence of temporal distance between frames.}
 \label{tab:temp_dist_table}
\end{table}

\newpage
\subsection{Late vs. Early Slow Fusion}
The results from Table~\ref{tab:temp_number_frames} show that albeit there is no obviously better architecture with respect to all experiments, the picture changes when regarding only the architectures that use an input sequence length of 4.
Considering only these architectures, the Late Slow Fusion clearly outperforms the Early Slow Fusion.

\begin{table}
    \centering
    \begin{tabular}{lcc@{ }c@{ }c}
        \toprule
		Architecture & \#Images &  $\Delta$AP Easy &$\Delta$AP Medium & $\Delta$AP Hard\\
		\midrule
		\textbf{LSF} & 4 & \textbf{15} & \textbf{13} & \textbf{14}  \\
		ESF          & 4 & 10 & 9 & 11 \\
		\bottomrule 
	\end{tabular}
\caption{Performance gains of Late Slow Fusion (LSF) and Early Slow Fusion (ESF) in relation to same networks without temporal information.}
 \label{tab:early_vs_late}
\end{table}

Additionally, both fusion architectures have been compared to their counterparts receiving no temporal information for different configurations. The relative performance gain of the fusion networks with respect to their non temporal counterparts are shown in Table~\ref{tab:early_vs_late}. Within this experiment, the Late Slow Fusion clearly outperforms the Early Slow Fusion on all KITTI difficulty levels.

\subsection{Temporal Padding}

\begin{table}
    \centering
    \begin{tabular}{lcc@{ }c@{ }c@{ }c}
        \toprule
		Architecture & 3D Blocks & \#Images &  AP Easy & AP Medium & AP Hard\\
		\midrule
		LSF & \textbf{Yes} & 4 & \textbf{0.939} & \textbf{0.910} & \textbf{0.830}  \\
		LSF & No  & 4 & 0.932 & 0.887 & 0.798 \\
		\bottomrule 
	\end{tabular}
\caption{Performance comparison for the Late Slow Fusion (LSF) with and without using 3D Dense Blocks for temporal padding.}
 \label{tab:temp_padding}
\end{table}
The influence of temporal padding on the LSF architecture is investigated by extending the dense blocks to 3D dense blocks and padding them in the temporal dimension. The resulting effect is determined by means of the performance change of the architectures. The results for this are illustrated in Table~\ref{tab:temp_padding}.
The results clearly show that temporal padding not only has no negative influence on the results, but even slightly better results are achieved in all KITTI difficulty levels.

\subsection{Temporal Fusion Blocks}

\begin{table}
    \centering
    \begin{tabular}{l@{ }cccc}
        \toprule
		Temporal Fusion & Arch & AP Easy & AP Medium & AP Hard\\
		\midrule
		$T \times 1 \times 1 $ Conv & LSF & 0.850 & 0.840 & 0.764  \\
		$T \times 3 \times 3 $ Conv & LSF & 0.809 & 0.818 & 0.745 \\
		$T \times 5 \times 5 $ Conv & LSF & 0.761 & 0.800 & 0.828 \\
		3D Inception v1             & LSF & 0.810 & 0.815 & 0.750 \\
		\textbf{3D Inception v2}    & LSF & \textbf{0.853} & \textbf{0.853} & \textbf{0.768} \\
		$T \times 3 \times 3 $ Max Pool & LSF & 0.194 & 0.345 & 0.376 \\
		$T \times 3 \times 3 $ Mean Pool & LSF & 0.168 & 0.329 & 0.359 \\
		\bottomrule 
	\end{tabular}
\caption{Performance of different temporal fusion blocks on the different difficulties within the KITTI dataset. Late Slow Fusion (LSF) architecture is used is this case.}
 \label{tab:temp_fusion_layer}
\end{table}

For determining the best temporal fusion block, a Late Slow Fusion architecture with an input sequence length of 4 images and a temporal distance of 400~ms between the images has been used. The results for all temporal fusion blocks are shown in Table~\ref{tab:temp_fusion_layer}.

The number of filters in the temporal layers was chosen in a way that the number of parameters in the temporal layers is constant. This series of tests was executed in parallel with the evaluation of temporal padding, which is why the architectures unfortunately do not yet feature temporal padding with 3D Dense Blocks. For this reason, the resulting numbers are not directly comparable with those from the other experiments.

The 3D Inception v2 architecture shows the best performance of all blocks, closely followed by the $T \times 1 \times 1$ convolution. Both blocks have a very low number of parameters, which results in the use of many filter kernels of the corresponding operations. A high spatial resolution seems to have a low impact on the overall performance, since the performance decreases as the spatial width of the kernel increases ($T \times 3 \times 3$ and $T \times 5 \times 5$). This may be due to the fact that a large number of convolutions have already been able to extract spatial information and few have been able to extract the temporal information. Since the Inception v2 block has better performance and is a more variable block, it is preferred over the $T \times 1 \times 1$ convolution.

\subsection{Temporal vs non-Temporal}

\begin{table}
    \centering
    \begin{tabular}{lc@{ }c@{ }c@{ }c}
        \toprule
		Architecture & Temporal Block & AP Easy & AP Medium & AP Hard\\
		\midrule
		LSF & 3D Inception v2  & \textbf{0.902} & \textbf{0.891} & \textbf{0.824}  \\
		2D & 2D Inception v2 & 0.876 & 0.868 & 0.792 \\
	    \midrule
		LSF & 3D Max Pooling  & \textbf{0.939} & 0.910 & \textbf{0.830}  \\
		2D & 2D Max Pooling   & 0.913 & \textbf{0.914} & 0.824 \\
		\bottomrule 
	\end{tabular}
\caption{Final performance comparison of Late Slow Fusion (LSF) and a non-temporal 2D architecture.}
 \label{tab:final_results}
\end{table}

For the final evaluation of our Temporal Feature Network, we selected the best performing components and compared them to a baseline model with almost the same design choices. 
Only the components containing temporal information, namely the 3D Inception and the 3D Dense Blocks are reduced to their non-temporal 2D versions.
In addition to the best performing temporal fusion blocks according to Table~\ref{tab:temp_fusion_layer} we added the 3D Max Pooling as it received surprisingly good results in combination with the 3D Dense blocks. 
The final results are shown in Table~\ref{tab:final_results}. Both, the 3D Inception v2 based network as well as the 3D Max Pooling network outperformed their non-temporal counterparts while the 3D Max Pooling in combination with 3D Dense blocks also outperformed the 3D Inception v2 network.
This confirms the results, which were already indicated in Table~\ref{tab:early_vs_late}.

\section{CONCLUSION}
With Temporal Feature Networks, in this work we presented a CNN architecture for extracting temporal information out of image sequences in order to enhance the detection performance of image based object detectors. An extensive series of experiments has  been conducted to find the optimal choice for each architectural component. 
Finally, a Late Slow Fusion network based on 3D Dense blocks and 3D Max Pooling blocks for temporal reduction performed best in combination of a input sequence size of 4 images with a temporal distance of 400 ms. This network clearly outperformed its non-temporal counterpart.

In general, our experiments have demonstrated that the use of temporal information enhances the performance of image based object detectors.
In a further step, it would be interesting to transfer these promising results to other perception tasks. Also the integration of Temporal Feature Networks into MultiTask networks might be an interesting option.








\bibliographystyle{IEEEtran}
\bibliography{IEEEabrv,main}

\begin{thebibliography}{10}
\providecommand{\url}[1]{#1}
\csname url@samestyle\endcsname
\providecommand{\newblock}{\relax}
\providecommand{\bibinfo}[2]{#2}
\providecommand{\BIBentrySTDinterwordspacing}{\spaceskip=0pt\relax}
\providecommand{\BIBentryALTinterwordstretchfactor}{4}
\providecommand{\BIBentryALTinterwordspacing}{\spaceskip=\fontdimen2\font plus
\BIBentryALTinterwordstretchfactor\fontdimen3\font minus
  \fontdimen4\font\relax}
\providecommand{\BIBforeignlanguage}[2]{{%
\expandafter\ifx\csname l@#1\endcsname\relax
\typeout{** WARNING: IEEEtran.bst: No hyphenation pattern has been}%
\typeout{** loaded for the language `#1'. Using the pattern for}%
\typeout{** the default language instead.}%
\else
\language=\csname l@#1\endcsname
\fi
#2}}
\providecommand{\BIBdecl}{\relax}
\BIBdecl

\bibitem{weber18}
M.~Weber, M.~Huber, and J.~M. Z{\"o}llner, ``{HDTLR: A CNN based Hierarchical
  Detector for Traffic Lights},'' in \emph{International Conference on
  Intelligent Transportation Systems (ITSC)}.\hskip 1em plus 0.5em minus
  0.4em\relax IEEE, 2018.

\bibitem{weber17}
M.~Weber, C.~Rist, and J.~M. Z{\"o}llner, ``{Learning Temporal Features with
  CNNs for Monocular Visual Ego Motion Estimation},'' in \emph{International
  Conference on Intelligent Transportation Systems (ITSC)}.\hskip 1em plus
  0.5em minus 0.4em\relax IEEE, 2017.

\bibitem{hubschneider17}
C.~Hubschneider, A.~Bauer, J.~Doll, M.~Weber, S.~Klemm \emph{et~al.},
  ``{Integrating End-to-End Learned Steering into Probabilistic Autonomous
  Driving},'' in \emph{International Conference on Intelligent Transportation
  Systems (ITSC)}.\hskip 1em plus 0.5em minus 0.4em\relax IEEE, 2017.

\bibitem{lin17a}
T.-Y. Lin, P.~Goyal, R.~Girshick, K.~He, and P.~Dollar, ``{Focal Loss for Dense
  Object Detection},'' in \emph{International Conference on Computer Vision
  (ICCV)}.\hskip 1em plus 0.5em minus 0.4em\relax IEEE, 2017.

\bibitem{he16}
K.~He, X.~Zhang, S.~Ren, and J.~Sun, ``{Deep Residual Learning for Image
  Recognition},'' in \emph{Conference on Computer Vision and Pattern
  Recognition (CVPR)}.\hskip 1em plus 0.5em minus 0.4em\relax IEEE, 2016.

\bibitem{liu16}
W.~Liu, D.~Anguelov, D.~Erhan, C.~Szegedy, S.~Reed, C.-Y. Fu, and A.~C. Berg,
  ``{SSD: Single Shot MultiBox Detector},'' in \emph{Advances in Neural
  Information Processing Systems (NIPS)}, 2016.

\bibitem{simonyan14}
K.~Simonyan and A.~Zisserman, ``{Very Deep Convolutional Networks for
  Large-Scale Image Recognition},'' in \emph{International Conference on
  Learning Representations (ICLR)}, 2015.

\bibitem{redmon17}
J.~Redmon and A.~Farhadi, ``{YOLO9000: Better, Faster, Stronger},'' in
  \emph{Conference on Computer Vision and Pattern Recognition (CVPR)}.\hskip
  1em plus 0.5em minus 0.4em\relax IEEE, 2017.

\bibitem{szegedy2016rethinking}
C.~Szegedy, V.~Vanhoucke, S.~Ioffe, J.~Shlens, and Z.~Wojna, ``{Rethinking the
  Inception Architecture for Computer Vision},'' in \emph{Conference on
  Computer Vision and Pattern Recognition (CVPR)}.\hskip 1em plus 0.5em minus
  0.4em\relax IEEE, 2016.

\bibitem{deng09}
J.~Deng, W.~Dong, R.~Socher, L.-J. Li, K.~Li, and L.~Fei-Fei, ``{ImageNet: A
  Large-Scale Hierarchical Image Database},'' in \emph{Conference on Computer
  Vision and Pattern Recognition (CVPR)}.\hskip 1em plus 0.5em minus
  0.4em\relax IEEE, 2009.

\bibitem{karpathy2014large}
A.~Karpathy, G.~Toderici, S.~Shetty, T.~Leung, R.~Sukthankar, and L.~Fei-Fei,
  ``{Large-Scale Video Classification with Convolutional Neural Networks},'' in
  \emph{Conference on Computer Vision and Pattern Recognition (CVPR)}, 2014.

\bibitem{Shen_2017_ICCV}
Z.~Shen, Z.~Liu, J.~Li, Y.-G. Jiang, Y.~Chen, and X.~Xue, ``{DSOD: Learning
  Deeply Supervised Object Detectors From Scratch},'' in \emph{International
  Conference on Computer Vision (ICCV)}.\hskip 1em plus 0.5em minus 0.4em\relax
  IEEE, 2017.

\bibitem{girshick14}
R.~Girshick, J.~Donahue, T.~Darrell, and J.~Malik, ``{Rich feature hierarchies
  for accurate object detection and semantic segmentation},'' in
  \emph{Conference on Computer Vision and Pattern Recognition (CVPR)}.\hskip
  1em plus 0.5em minus 0.4em\relax IEEE, 2014.

\bibitem{redmon16}
J.~Redmon, S.~Divvala, R.~Girshick, and A.~Farhadi, ``{You Only Look Once:
  Unified, Real-Time Object Detection},'' in \emph{Conference on Computer
  Vision and Pattern Recognition (CVPR)}.\hskip 1em plus 0.5em minus
  0.4em\relax IEEE, 2016.

\bibitem{bochkovskiy20}
A.~Bochkovskiy, C.-Y. Wang, and H.-Y.~M. Liao, ``{YOLOv4: Optimal Speed and
  Accuracy of Object Detection},'' \emph{CoRR}, vol. abs/2004.10934, 2020.

\bibitem{redmon18b}
J.~Redmon and A.~Farhadi, ``{YOLOv3: An Incremental Improvement},''
  \emph{CoRR}, vol. abs/1804.02767, 2018.

\bibitem{sermanet13}
P.~Sermanet, D.~Eigen, X.~Zhang, M.~Mathieu, R.~Fergus, and Y.~LeCun,
  ``{OverFeat: Integrated Recognition, Localization and Detection using
  Convolutional Networks},'' \emph{CoRR}, vol. abs/1312.6229, 2013.

\bibitem{huval15}
B.~Huval, T.~Wang, S.~Tandon, J.~Kiske, W.~Song, J.~Pazhayampallil
  \emph{et~al.}, ``{An Empirical Evaluation of Deep Learning on Highway
  Driving},'' \emph{CoRR}, vol. abs/1504.01716, 2015.

\bibitem{weber19}
M.~Weber, M.~{Fürst}, and J.~M. Z{\"o}llner, ``{Direct 3D Detection of
  Vehicles in Monocular Images with a CNN based 3D Decoder},'' in
  \emph{Intelligent Vehicles Symposium (IV)}.\hskip 1em plus 0.5em minus
  0.4em\relax IEEE, 2019.

\bibitem{gahlert19}
N.~G{\"a}hlert, J.-J. Wan, M.~Weber, J.~M. Z{\"o}llner, U.~Franke, and
  J.~Denzler, ``{Beyond Bounding Boxes: Using Bounding Shapes for Real-Time 3D
  Vehicle Detection from Monocular RGB Images},'' in \emph{Intelligent Vehicles
  Symposium (IV)}.\hskip 1em plus 0.5em minus 0.4em\relax IEEE, 2019.

\bibitem{jegou2017one}
S.~J{\'e}gou, M.~Drozdzal, D.~Vazquez, A.~Romero, and Y.~Bengio, ``{The One
  Hundred Layers Tiramisu: Fully Convolutional DenseNets for Semantic
  Segmentation},'' in \emph{Conference on Computer Vision and Pattern
  Recognition (CVPR)}.\hskip 1em plus 0.5em minus 0.4em\relax IEEE, 2017.

\bibitem{huang2017densely}
G.~Huang, Z.~Liu, L.~Van Der~Maaten, and K.~Q. Weinberger, ``{Densely Connected
  Convolutional Networks},'' in \emph{Conference on Computer Vision and Pattern
  Recognition (CVPR)}.\hskip 1em plus 0.5em minus 0.4em\relax IEEE, 2017.

\bibitem{zhu2017flow}
X.~Zhu, Y.~Wang, J.~Dai, L.~Yuan, and Y.~Wei, ``{Flow-Guided Feature
  Aggregation for Video Object Detection},'' in \emph{International Conference
  on Computer Vision (ICCV)}.\hskip 1em plus 0.5em minus 0.4em\relax IEEE,
  2017.

\bibitem{simonyan14b}
K.~Simonyan and A.~Zisserman, ``{Two-Stream Convolutional Networks for Action
  Recognition in Videos},'' in \emph{Advances in Neural Information Processing
  Systems (NIPS)}, 2014.

\bibitem{geiger12}
A.~Geiger, P.~Lenz, and R.~Urtasun, ``{Are we ready for Autonomous Driving? The
  KITTI Vision Benchmark Suite},'' in \emph{Conference on Computer Vision and
  Pattern Recognition (CVPR)}.\hskip 1em plus 0.5em minus 0.4em\relax IEEE,
  2012.

\bibitem{romera18}
E.~Romera, L.~M. Bergasa, J.~M. Alvarez, and M.~Trivedi, ``{Train Here, Deploy
  There: Robust Segmentation in Unseen Domains},'' in \emph{Intelligent
  Vehicles Symposium (IV)}.\hskip 1em plus 0.5em minus 0.4em\relax IEEE, 2018.

\bibitem{kingma2014adam}
D.~P. Kingma and J.~Ba, ``{Adam: A Method for Stochastic Optimization},''
  \emph{International Conference on Learning Representations (ICLR)}, 2015.

\bibitem{everingham10}
M.~Everingham, L.~Van~Gool, C.~K.~I. Williams, J.~Winn, and A.~Zisserman,
  ``{The PASCAL Visual Object Classes (VOC) Challenge},'' \emph{International
  Journal of Computer Vision}, vol.~88, no.~2, pp. 303--338, 2010.

\end{thebibliography}

\end{document}